# An Augmented Transformer Architecture for Natural Language Generation Tasks

Hailiang Li[1], Adele, Y.C. Wang[1], Yang Liu[1], Du Tang[1], Zhibin Lei[1], Wenye Li[2]
[1]Hong Kong Applied Science and Technology Research Institute Company Limited, Hong Kong, China
Email: {harleyli, adelewang, yangliu, tangdu, lei}@astri.org
[2]The Chinese University of Hong Kong, Shenzhen, China
Email: wyli@cuhk.edu.cn

*Abstract*—The Transformer based neural networks have been showing significant advantages on most evaluations of various natural language processing and other sequence-to-sequence tasks due to its inherent architecture based superiorities. Although the main architecture of the Transformer has been continuously being explored, little attention was paid to the positional encoding module. In this paper, we enhance the sinusoidal positional encoding algorithm by maximizing the variances between encoded consecutive positions to obtain additional promotion. Furthermore, we propose an augmented Transformer architecture encoded with additional linguistic knowledge, such as the Part-of-Speech (POS) tagging, to boost the performance on some natural language generation tasks, e.g., the automatic translation and summarization tasks. Experiments show that the proposed architecture attains constantly superior results compared to the vanilla Transformer.

## I. INTRODUCTION

Previously, the dominant approaches to natural language generation (NLG) tasks were heavily dependent on recurrent neural networks (RNNs) [32] to model the temporal dynamics in the sequence data. Although most RNN structures are well-designed, they still suffer from long-term dependency problems and low training-inference speed. The Transformer architecture [30] is proposed to mitigate these defects. It sets up a recipe that only fully connected layer and dot-product attention mechanisms are used, and entirely circumvent all the recurrence.

The block diagram of the Transformer is described briefly in Fig. 1, where the grey part denoted the whole block diagram except two dash line blocks illustrates the encoder-decoder architecture based the vanilla Transformer [30]. Two grey blocks denote the Transformer encoder and decoder comprising multi-head attention layers followed by position-wise feed forward layers, residual connections [12] and layer normalization [4].

These days, while most of researchers are continually ameliorating the main components of the Transformer architecture, i.e., the Transformer encoder and decoder components, less attention was paid to the positional encoding parts (the orange circles in Fig. 1). In this work, we refined the existing sinusoidal positional encoding (PE) algorithm by maximizing all the variances between encoded consecutive signals (positions), namely maximum variances Positional Encoding (mvPE), which can obtain additional promotion in some NLG tasks. Moreover, we employed the sinusoidal positional encoding to inject additional linguistic knowledge into input and output embeddings. Precisely, we encoded the Part-of-Speech (POS) tagging [6, 34] (as shown in the two dash line blocks in Fig. 1) for proposed augmented Transformer to boost the performance of some natural language generation tasks, e.g., the translation and the summarization.

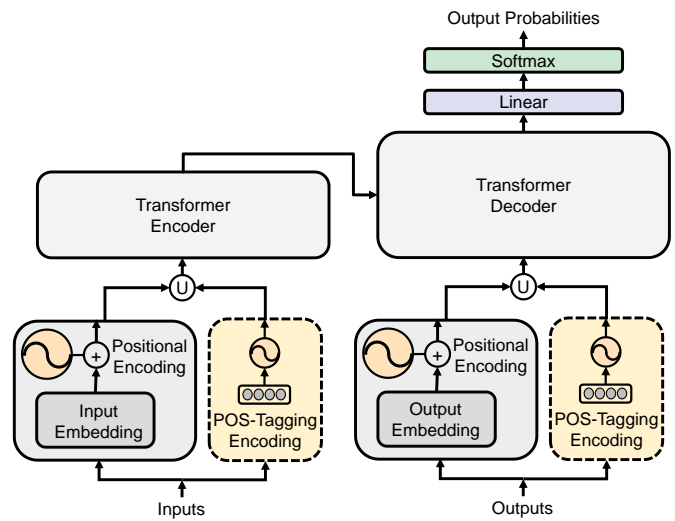

Fig. 1: The block diagram of the proposed augmented Transformer model architecture with the refined positional encoding algorithm and the additional POS tagging encoding module.

The core contributions of this work are summarized as follows: (1) We demonstrated an improved sinusoidal positional encoding method by maximizing the variances between encoded consecutive signals; (2) We proved that the sinusoidal positional encoding module can further incorporate extra language priors, such as POS tagging information, into the Transformer based models, which derived an augmented Transformer; (3) We evaluated the augmented Transformer on some NLG tasks, e.g., translation, on dataset with modest scale (1 million samples), and results show the new model achieves value-added improvement comparing to the vanilla Transformer.

## II. RELATED WORK

### A. From Vanilla RNN to the Transformer

The sequence to sequence (S2S) based natural language processing (NLP) tasks are conducted by the deep learning models with an encoder-decoder manner. Encoders are trained to represent the contextual information within input sequence with encodings, while decoders are trained to map these encodings to the target sequences.

RNNs [32], in particularly the Long Short-Term Memory (LSTM) networks [13] and the Gated Recurrent Units (GRUs) [24], have been essential components of lots of NLP models previously, and they also have showed outstanding performance on lots of benchmarks.

Although RNNs are widely adopted, there are still some drawbacks on RNNs: Back propagation procedure through time usually suffers from gradient vanishing and explosion [14]; The training process is also hard to be parallelized because of the consecutive operations. What's more, previous work has showed that LSTM is hard to tackle with longer sentences (say, more than 200 context words) [15], which hinders their further spreading.

Recently, attention mechanisms have paved a path to infer the direct connections between long-distance word pairs, e.g., the intra-sentence attention [9], the recurrent additive networks [18], and the Transformer architecture [30].

The framework of an attention based model generally consists of two RNNs: *Encoder* and *Attention-Decoder*, which is capable of tackling two sequences with different lengths based on the cross-entropy criterion. The *Encoder* module tries to encode the input vectors, $\boldsymbol{x} = (x_1, x_2, \dots x_L)$ into high-level representation vectors: $\boldsymbol{h} = (h_1, h_2, \dots h_M)$. Then, the Attention-Decoder produces the output in a form of probability distribution over labels $\boldsymbol{y} = (y_1, y_2, \dots y_N)$ conditioned on the hidden context $\boldsymbol{h}$ and all the previous seen labels: $y_{1:n-1}$ according to the recursive Eqn. (1), where $L$ is the length of input vector, $M$ is the hidden dimension size, and $N$ is the number of district labels.

$$P(\boldsymbol{y}|\boldsymbol{x}) = \prod_n P(y_n|\boldsymbol{x}, y_{1:n-1}) \quad (1)$$

$$\boldsymbol{h} = Encoder(\boldsymbol{x}) \quad (2)$$

$$y_n \sim AttentionDecoder(\boldsymbol{h}, y_{1:n-1}) \quad (3)$$

Therefore, the loss function of the attention mechanism based encoder-decoder transduction model can be computed from following:

$$\mathcal{L}_{Attention} = -\sum_n \ln P(y_n|\boldsymbol{x}, y_{1:n-1}) \quad (4)$$

The dot-product attention based models are capable of alleviating the long-term dependency problem and preventing the vanishing gradient problem [14]. Meanwhile, the self-attention mechanism [30] can further promote the parallel computation, which derives the Transformer architecture.

Attention mechanisms, particularly the intra-inter attention modules based Transformer architecture [30] has shown significant advances in lots of NLP benchmarks. Nowadays, The Transformer architecture has built firm foundation for transfer learning in NLP tasks. Lots of unsupervised pretraining models based on Transformer have emerged: the generative pre-training (GPT) model [28], the AllenNLP research library [10], the Bidirectional Encoder Representation from Transformers (BERT) model [22], and other sequence-to-sequence tasks, such as image captioning [37, 38] and video description [35, 36]. So, to further improve the performance of Transformer becomes a critical mission in NLP and relevant domains.

The vanilla Transformer based model consists two parts in its input: the word embeddings and the position embeddings (see Fig. 1). The word embedding can be generated by an unsupervised method such as widely used methods CBOW [25] and GloVe [26]. Meanwhile, the position embedding part can also be explored for encoding language priors.

### B. The Positional Encoding in Transformer

To ensure that the Transformer model can explicitly use the order of the sequence, some information about the relative or absolute positions of the tokens in the sequence should be injected into the input encoding vectors. In a Transformer based architecture, the positional encoding modules (denoted as the orange circles in Fig. 1), are used to deliberately encode the relative/absolute positions of the inputs as vectors. The positional encoding modules generates vectors of the same dimension with model size $d_{model}$. It is merged to the input embedding vectors by element-wise addition operation to keep the dimension size unchanged.

There are many choices of positional encoding, while in the Transformer architecture, sine and cosine functions of different frequencies are used for encoding corresponding even and odd positions as shown in Eqn. (5) and Eqn. (6) respectively:

$$PE(pos, 2i) = sin(\frac{pos}{max\_length^{\frac{2i}{d_{model}}}}) \quad (5)$$

$$PE(pos, 2i+1) = cos(\frac{pos}{max\_length^{\frac{2i}{d_{model}}}}) \quad (6)$$

where $pos$ is the position and $i$ is the index value. Each dimension of the positional encoding corresponds to a sinusoidal point, and each position is mapped to a sinusoidal curve as shown in Fig. 2. The wavelengths form a geometric progression ranged from $2\pi$ to $max_{length} \cdot 2\pi$, ($max\_length$ stands for the max length of encoding tokens, e.g., 500). This sinusoidal function is chosen because it would allow the model to easily encode the relative positions, since any fixed offset $k$, $PE(pos + k)$ can be represented as a linear function of $PE(pos)$ as demonstrated in following:

$$\begin{bmatrix} \sin(pos + k) \\ \cos(pos + k) \\ \dots \end{bmatrix} = \begin{bmatrix} \sin(pos)\cos(k) + \cos(pos)\sin(k) \\ \cos(pos)\cos(k) - \sin(pos)\sin(k) \\ \dots \end{bmatrix} \quad (7)$$

## III. A REFINED POSITIONAL ENCODING

It has been approved that the delicately designed Positional Encoding (PE) module [30] can work efficiently for unlimited positions theoretically. However, little attention was paid to the PE module effectiveness and extensibility. In a recent work [31] on this research, authors present an approach to represent relative positions, rather than absolute positions. In order to analyze the work scheme of the PE module and to explore its effectiveness, we plotted the sinusoidal positional encoding curves (as shown in Fig. 2) and made some comparisons between consecutive curves, which ignited us some enlightenment later.

### A. Weakness of Positional Encoding

After observing the plotted consecutive sinusoidal curves, as part of them shown in Fig. 2 (curves of position 1 and 2, and curves of position 99 and 100), we can intuitively bring forth a conclusion: the variances between consecutive positions in the front part, e.g., position 1 and 2, can get larger values, while those variances between consecutive positions in the end part, e.g., position 99 and 100, will get relatively smaller values. Theoretically, in signal processing, the smaller variances between encoded signals will weaken the representation of a signal system, consequently, we did experiments and verified our hypothesis. Therefore, in order to improve the signal representation capability of the positional encoding module, a reasonable solution is to maximize the variances between consecutive encoded positions, particularly for positions in the end part of the tokens in a sequence.

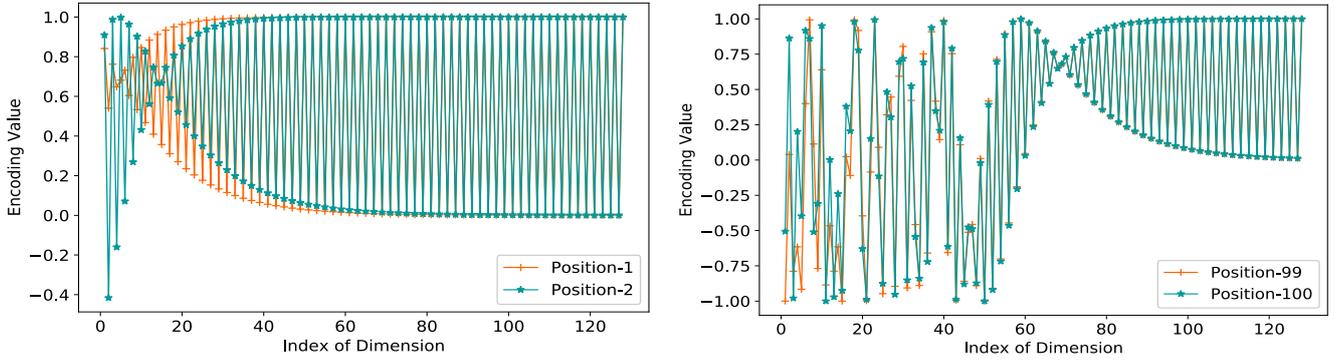

Fig. 2: The sinusoidal positional encoding curves (dimension: 128) with different positions (curves of position 1 and 2 are on the left part, and curves of position 99 and 100 are on the right part).

### B. The Maximum Variances Positional Encoding Algorithm

In this paper, a simple yet effective approach is proposed to maximize the variances between encoded consecutive signals (positions). We consider a solution with a step parameter, $k \geq 1$, in the Positional Encoding ($PE$), e.g., Position 1 is encoded as original $PE(k)$, Position 2 is encoded as original $PE(2k)$, ..., and so on. It is clear that the vanilla $PE$ algorithm is the solution with step: $k = 1$. So, the best solution is the one with an optimal step $k$. Based on the object to maximize the variances between encoded consecutive signals, a target function can be defined as Eqn. (8):

$$\arg\max_k (\sum_{\substack{1 \leq i < j < L \\ 1 \leq k}} |PE(i*k) - PE(j*k)|) \quad (8)$$

where $L$ is the max length of the input, e.g., the max sentence length on NLP tasks.

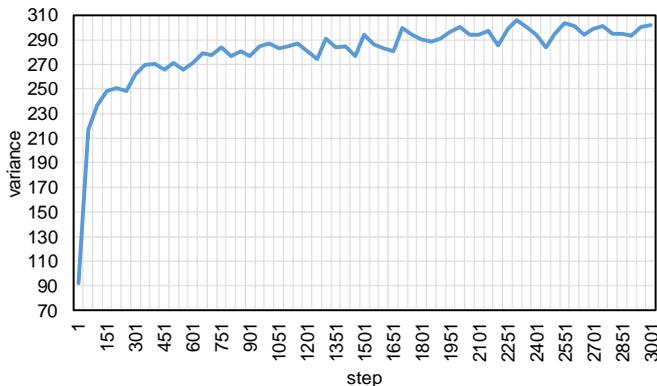

Fig. 3: The total (summed) variance between encoded signals by the positional encoding (PE) algorithm according to different steps.

When the variance values with different steps (parameters) are visualized in Fig. 3, we can grasp an intuitive understanding where an optimal value of step $k$ can be obtained with respect to the variance. As shown in Fig. 3, in the front part within 1000 steps, the variance gets larger when step $k$ increases, later the curve comes to a plateau with little fluctuation. Therefore, an optimal value can be chosen in the plateau area. Once the optimal step $k$ has obtained, the positional encoding values can be calculated with the following new formulas, Eqn. (9) and Eqn. (10). Based on the described strategy, we named the proposed PE algorithm as maximum variances Positional Encoding (mvPE).

$$PE(pos, 2i) = sin(\frac{pos*k}{max\_length^{\frac{2i}{d_{model}}}}) \quad (9)$$

$$PE(pos, 2i+1) = cos(\frac{pos*k}{max\_length^{\frac{2i}{d_{model}}}}) \quad (10)$$

## IV. A NOVEL TRANSFORMER

After refining the PE module, a natural idea arises: Is the PE or mvPE algorithm able to be extended to encode other discrete features as additional language priors in Transformer based models?

### A. Prior Knowledge for Deep Learning

For decades, most research works on machine learning are feature engineering. Deep learning based networks, such as the CNN (convolutional neural networks) [17] network can automatically implement the feature engineering work by stacking lots of convolutional filters. Meanwhile, RNN [32] and attention based networks, such as the Transformer model, are lack of this scheme explicitly. On the NLP tasks, word embedding pre-processing is the prior extracting technique, as it is capable of projecting some discrete language prior information, such as sentence syntax and semantic relations, into the real-number space, thus it is always a fundamental research topic in NLP domain.

### B. POS Tagging as Language Prior

To tackle the issue of encoding more linguistic knowledge into current NLP networks, particularly the Transformer architecture, language prior encoding needs to be explored.

When analyzing a sentence, the main sentence structure, i.e., the Subject–Verb–Object (SVO) components, need be paid with more attention. The sentence syntax and semantic can be an aid on the NLP tasks, which has been exploring by linguists and other researchers. Though syntax and semantic analysis can assist NLP tasks, it is hard to find a stable algorithm to inject the extra information into models in most NLP tasks. We find an alternative solution, which can achieve a similar effect, the Part-of-Speech (POS) tagging. POS Tagging [6, 7, 8, 34] is the task to extract the correct POS tag for each word in one sentence, e.g., noun, verb, adverb, and so on. There are not too many POS tags for one language. We choose the Penn Treebank POS tag [19] set containing only 36 POS tags (as briefly described in TABLE I) for our experiments in the English corpora. And we use the tag mapping with only 40 POS tags used in the Chinese words' segmentation toolkit (jieba: https://pypi.org/project/jieba) in our Chinese corpora.

| No | Tag | Description |
|---|---|---|
| 1 | CC | Coordinating conjunction |
| 2 | CD | Cardinal number |
| … … | | |
| 12 | NN | Noun, singular or mass |
| … … | | |
| 27 | VB | Verb, base form |
| … … | | |
| 35 | WP$ | Possessive wh-pronoun |
| 36 | WRB | Wh-adverb |

TABLE I: list of part-of-speech (POS) tags.

POS Tagging is widely used for information extraction and retrieval, and speech processing. Currently, a lot of tools are available to do POS Tagging efficiently. Moreover, POS tag of some ambiguous words can be detected correctly. For example, the word "color" can be treated as a noun or a verb in the sentence: "use the red color to color the book cover.", which can be tagged correctly by toolkit (https://www.nltk.org) as shown in TABLE II.

| POS tagging Example |
|---|
| **Encoding Sentence:** |
| text = "use the red **color** to **color** the book cover ." |
| **Result:** |
| ('use', 'VB'), ('the', 'DT'), ('red', 'JJ'), (**'color'**, **'NN'**), ('to', 'TO'), (**'color'**, **'VB'**), ('the', 'DT'), ('book', 'NN'), ('cover', 'NN'), ('.', '.') |

TABLE II: An example of the POS tagging result on one sentence with the toolkit: nltk.

### C. Augmented Transformer Architecture

The POS tagging information for each word is like the position of each word, and they can also be represented as integer values. Same as the position values, the POS tagging information can be encoded by the PE or mvPE module. In the original PE scheme, the PE encoded vector is added to the input word embedding vector. The refined mvPE will use concatenating, instead of summing, to encode the POS tagging, as shown in Fig. 1. The POS tagging information is encoded into a vector of relatively smaller size to balance the computation. We set the dimension size to be 64 in our experiments.

Based on the idea described above, we propose an augmented Transformer architecture (as shown in Fig. 1) for the NLG tasks. The input vectors for the encoder and decoder in the Transformer are generated with the following formula:

$$V_{emb} \oplus V_{mvPE} \cup V_{PoST} \quad (11)$$

where vector $V_{emb}$ is the input word embedding vector with dimension $d_{emb}$, $V_{mvPE}$ is the vector coming from the proposed mvPE with dimension $d_{mvPE}$ (equals to $d_{emb}$), and $V_{PoST}$ is the vector with dimension $d_{PoST}$ generated by POS tagging encoding module. The additional POS tagging encoding module can work with ignorable overload, and the dimension of $d_{emb}$ and $d_{PoST}$ can set as proper values. In our experiments, $d_{PoST}$=64, and $d_{emb}$=$d_{mvPE}$=300.

We conducted some comparisons among the vanilla Transformer, the Transformer with mvPE module and our proposed augmented Transformer model with POS tagging module inside in two NLG tasks: English-Chinese translation and English sentence abstractive summarization [1, 33]. Both show a similar tendency on the test perplexity performance, and we choose the translation to report in detail to avoid the duplications of similar figures.

As shown in Fig. 4, the Transformer with refined mvPE module can achieve more than 1~2% improvement on test perplexity metrics in comparison to the vanilla Transformer model; meanwhile, the augmented Transformer with POS tagging encoding module inside can obtain more than 5% improvement on average.

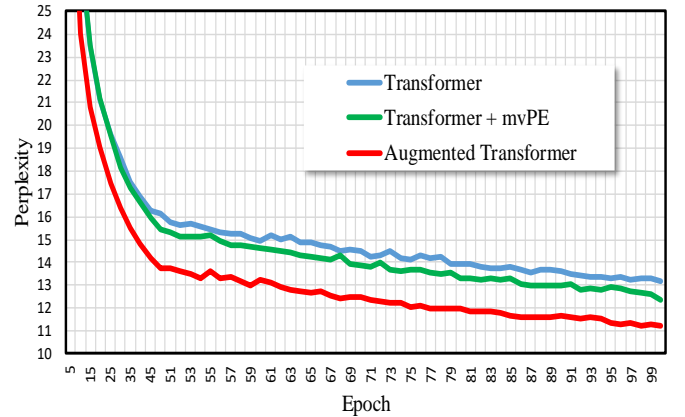

Fig. 4: Test perplexities of: the vanilla Transformer model, the Transformer with the maximum variances Positional Encoding (mvPE) module, and the augmented Transformer with POS tagging encoding module inside.

### V. EXPERIMENTS

We conducted experiments for these three models on the English-Chinese translation, and we give detailed analysis in the following sub-sections. As the refined Positional Encoding: mvPE and the proposed POS tagging encoding module in the augmented Transformer are additional modules to the vanilla Transformer architecture, most of the comparisons are carried on among two proposed models and the vanilla Transformer model.

Our analysis leverages these models on variety of machine translation automatic evaluation metrics, including the BLEU (bilingual evaluation understudy) [27] score, the ROUGE (Recall-Oriented Understudy for Gisting Evaluation) [20] score, the METEOR (Metric for Evaluation of Translation with Explicit ORdering) [2] score, and the F1 score [5]. Among all the matrices, the BLEU and the ROUGE scores are the primary evaluation matrices.

## A. Neural Machine Translation

Neural machine translation (NMT) has achieved the state-of-the-art (SoTA) performance on language translation tasks. These SoTA algorithms include the RNNs, firstly introduced Long Short-Term Memory (LSTM) based variants [29], the CNNs, proposed by [11], and the recently announced Transformer architecture [30]. A typical attention based NMT model is an end-to-end neural network framework, which consists of an encoder, a decoder and an attention mechanism [3]. During the training process, the NMT models try to explore the relations from a source language to a target language using the fed parallel bilingual corpus.

For general NMT algorithms, most NMT classic work [3, 21, 23] need be compared on some benchmark datasets. However, since our proposed algorithm is based recently Transformer architecture and holds the addition property, only Transformer based models are put in the comparison list. In our work, the English-Chinese translation is conducted on the classic Transformer model and two proposed models, and comparisons are illustrated on the three models.

## B. Dataset and Training Setting

Our dataset comes from the recently released large scale open dataset for English-Chinese Machine Translation Competition in the 2018 Challenger AI (artificial intelligence) event (https://challenger.ai/competition/ect2018). We make the same setting for all the models, with one million samples for training the models and 50,000 samples for evaluating the trained models.

Recently, some NMT experimental results and algorithms can be carried out using some open source projects, and openNMT is such an easily operated tool focuses on NMT (https://github.com/OpenNMT/OpenNMT) [16]. We also have developed our augmented model based on the python code from openNMT. In all experiments, we set the word embedding dimension size as 300, hidden units as 512, encoder and decoder depth as 4, batch size as 1024 with all the models. To get reliable yet useful models for testing, all the models are trained for 80,000 steps (around 12 hours) in an Intel x64 desktop machine with i7-7700K CPU @4.20GHz, and with a GeForce GTX 1080-Ti Graphics Card.

## C. Evaluation with Metrics

These days, an opened tool for holistic analysis of language generations systems, such as: translation and summarization, can be employed (https://github.com/neulab/compare-mt) to derive detailed validation reports for evaluated models.

|  | Transformer | mvPE | Augmented |
|---|---|---|---|
| **BLEU-1** | 0.4613 | 0.4563 | 0.4619 |
| **BLEU-2** | 0.2808 | 0.2826 | 0.2861 |
| **BLEU-3** | 0.1846 | 0.1885 | 0.1907 |
| **BLEU-4** | 0.1259 | 0.1303 | 0.1312 |
| **METEOR** | 0.2273 | 0.2297 | 0.2328 |
| **ROUGE-L** | 0.4582 | 0.4644 | 0.4683 |

TABLE III: Evaluation scores on different metrics (as listed) for the three models (mvPE denotes the Transformer + mvPE and Augmented denotes the proposed augmented Transformer).

We evaluated all the three models with different objective evaluation metrics: BLEU, METEOR and ROUGE. As all the scores shown in TABLE III, these experimental results further prove that the Transformer + mvPE based model outperforms the baseline of the vanilla Transformer, except on the BLEU-1 metrics item. Moreover, TABLE III shows the augmented Transformer model's performance stably exceeds other two models in all the evaluation metrics. Meanwhile, on ROUGE-L criteria, both mvPE and the augmented model get value-added improvement compared to the vanilla Transformer model.

TABLE IV shows the detail measure scores in the ROUGE [20] metric, containing ROUGE-1, ROUGE-2 and ROUGE-L, in F1 measure [5] score (denoted as "F" in superscript), precision score (denoted as "P" in superscript) and recall score (denoted as "R" in superscript). As shown in TABLE IV, in all the three measure items, i.e., F1 measure, precision and recall, mvPE model stalely wins the baseline of the vanilla Transformer in all the ROUGE metrics. Meanwhile, the augmented Transformer model defeats the other two in all the measure items.

|  | Transformer | mvPE | Augmented |
|---|---|---|---|
| **ROUGE-1$^F$** | 0.4892 | 0.4932 | 0.4982 |
| **ROUGE-1$^P$** | 0.5194 | 0.5300 | 0.5387 |
| **ROUGE-1$^R$** | 0.4764 | 0.4754 | 0.5387 |
| **ROUGE-2$^F$** | 0.1880 | 0.1935 | 0.1968 |
| **ROUGE-2$^P$** | 0.1987 | 0.2069 | 0.2123 |
| **ROUGE-2$^R$** | 0.1843 | 0.1878 | 0.1896 |
| **ROUGE-L$^F$** | 0.4563 | 0.4606 | 0.4655 |
| **ROUGE-L$^P$** | 0.4984 | 0.5100 | 0.5192 |
| **ROUGE-L$^R$** | 0.4577 | 0.4582 | 0.4605 |

TABLE IV: Evaluation scores details on the ROUGE metric (mvPE denotes the Transformer + mvPE and Augmented denotes the proposed augmented Transformer).

TABLE V shows the output sentence length ratio for all the three models. It is clear both mvPE and the augmented Transformer models can get shorter translation sentences in comparison to the vanilla Transformer. TABLE V also denotes that the mvPE model gets more short sentences compare to the augmented Transformer model.

| Length Ratio | | |
|---|---|---|
| **Transformer** | **mvPE** | **Augmented** |
| 0.9884 | 0.9218 | 0.9440 |
| (ref=508319, out=502445) | (ref=508319, out=468555) | (ref=508319, out=479832) |

TABLE V: The output sentence length ratio from the three models (ref means the total reference sentence length and out means the total output sentence length).

## D. Word and Sentence Bucket Analysis

Bucket words or sentences by various statistics (e.g., word frequency, sentence BLEU, length difference with the reference, overall length), and calculate statistics by bucket (e.g., number of sentences, BLEU score per bucket) can give more details on analyzing the performance of proposed models. A number of measurements and analyses with figures are carried on in following, which attempt to pick out salient differences between the models, and make it easier to figure out what things one model is doing better than others.

The word-level F1-measure relating to word frequency analysis is to evaluate whether a model can do better at low-frequency words. Fig. 5 shows the measurement with the word-level F1-measure by frequency bucket. From word-level F1-measure by frequency bucket analysis, we can see that both the mvPE and the augmented Transformer models can outperform the vanilla Transformer model on working words with higher frequency. Meanwhile, the proposed two models seem to be better to tackle the words with lower frequency compare to the vanilla Transformer.

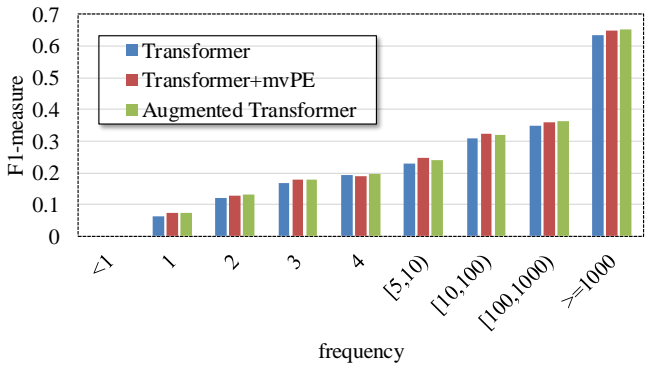

Fig. 5: Word-level F1-measure relating to frequency bucket.

The Sentence-level BLEU measure relating to sentence length analysis is to evaluate whether a model can do better at shorter sentences. From the sentence-level BLEU measure by length bucket analysis, as shown in Fig. 6, we can see the mvPE is good at short sentences but is poor at long sentences. On the other hand, the augmented Transformer can work well on almost all the short and long sentences compared to the other two models. Finally, in overall looking, all Transformer based models get acceptable results, since they all have the potential capability of capturing the long-term dependencies between sequences as scholars designing the Transformer architecture expected.

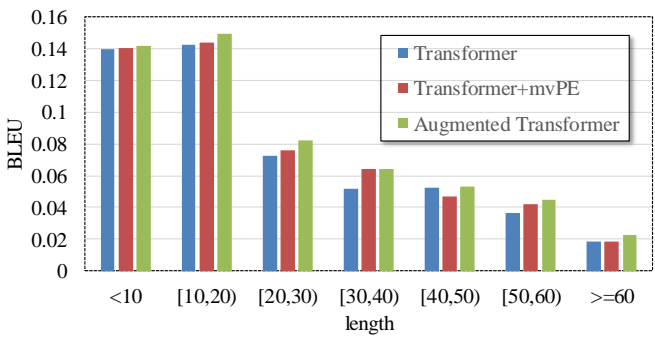

Fig. 6: Sentence-level BLEU relating to by length bucket.

The measure of count relating to the reference-output length difference attempts to evaluate the generated sentence concise of a model, which is helpful on summarization evaluation, but also useful in the translation evaluation task. According to the sentence count measure by reference-output length difference bucket analysis, as shown in Fig. 7, from the whole point of view, all the three models generate more short-sentences than long-sentences compared to the reference sentences. Relatively, the mvPE model gets more short sentences, than the augmented Transformer model, following by the baseline vanilla Transformer model. This result can match the phenomenon of the output sentence length ratio in TABLE V.

The measure of sentence count relating to the sentence-level BLEU attempts show the model performance in detail with buckets. As reported from Fig. 8, for all the three models, more than 70% sentences get the BLEU score less than 0.3, and around 30% sentences can achieve relatively higher score, i.e., larger than BLEU score 0.3. This result may be caused by insufficient training. At the same time, it is clear that both the mvPE model and the augmented Transformer model can get a relatively more sentences in the BLEU higher score buckets compared to the vanilla Transformer based model. Meanwhile, the augmented model works better than mvPE model in BLEU criteria.

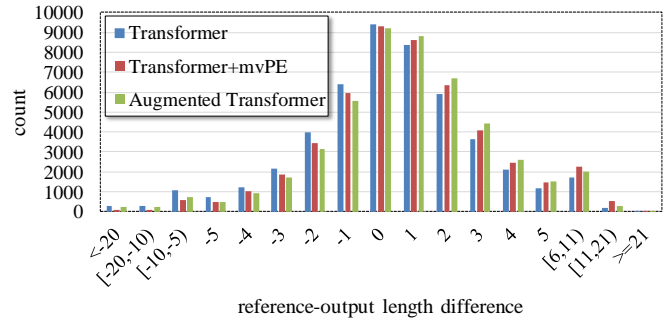

Fig. 7: Sentence count relating to reference-output length difference bucket.

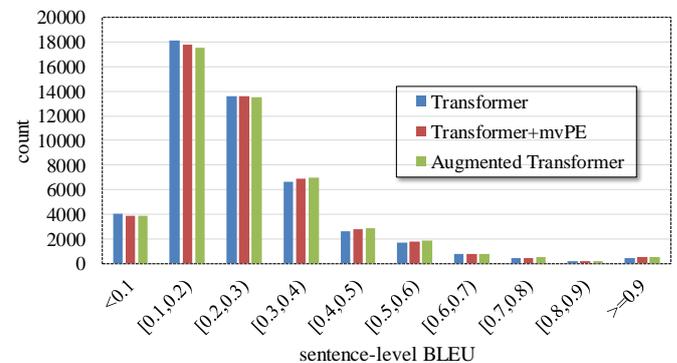

Fig. 8: Sentence count relating to sentence-level BLEU bucket

## VI. CONCLUSIONS

In this work, a number of contributions have been made to the Transformer architecture. Firstly, we addressed the problem of small variances in the existing Positional Encoding (PE) module of the vanilla Transformer, as well as refined the original PE component by maximizing the variances between encoded consecutive signals (positions), named as maximum variances Positional Encoding (mvPE). Meanwhile, the original PE can be explained as the primitive version of mvPE as step parameter: $k = 1$. Secondly, we explored the way to add linguistic knowledge for the Transformer, and proposed a novel augmented Transformer architecture which extends the PE/mvPE module to incorporate additional language priors, such as sentence Part-of-Speech (POS) tagging information. Finally, we evaluated proposed models on the translation task in a relatively low-resource setting, and our proposed models achieve stable improvement compared to the vanilla Transformer. Particularly, more than 5% improvement can be achieved on the augmented Transformer with the POS tagging encoding module inside. Future work includes evaluating proposed models with more natural language generation (NLG) tasks on large scale datasets, and exploring more language priors, as well as encoding them with the proposed augmented Transformer architecture.

## REFERENCES

[1] Abigail See, Peter J Liu, and Christopher D Manning. 2017. Get to the point: Summarization with pointer-generator networks, In *Proceedings of Association for Computational Linguistics (ACL),* 2017.

[2] Banerjee, S. and Lavie, A. 2005. METEOR: An Automatic Metric for MT Evaluation with Improved Correlation with Human Judgments, In *Proceedings of Workshop on Intrinsic and Extrinsic Evaluation Measures for MT and/or Summarization at the 43rd Annual Meeting of the Association of Computational Linguistics (ACL-2005).*

[3] Dzmitry Bahdanau, Kyunghyun Cho, and Yoshua Bengio. 2014. Neural machine translation by jointly learning to align and translate. arXiv preprint arXiv:1409.0473.

[4] J. L. Ba, J. R. Kiros, and G. E. Hinton. 2016. "Layer normalization," arXiv:1607.06450

[5] Beitzel, S.M., 2006. On understanding and classifying web queries (pp. 3216-3216). Chicago: Illinois Institute of Technology.

[6] Bernd Bohnet and Joakim Nivre. 2012. A TransitionBased System for Joint Part-of-Speech Tagging and Labeled Non-Projective Dependency Parsing. In *Proceedings of EMNLP-CoNLL.* pages 1455–1465.

[7] Chris Alberti, David Weiss, Greg Coppola, and Slav Petrov. 2015. Improved Transition-Based Parsing and Tagging with Neural Networks. In *Proceedings of EMNLP.* pages 1354–1359.

[8] Dat Quoc Nguyen, Mark Dras, and Mark Johnson. 2017. A Novel Neural Network Model for Joint POS Tagging and Graph-based Dependency Parsing. In *Proceedings of the CoNLL 2017 Shared Task.* pages 134–142.

[9] Jianpeng Cheng, Li Dong, and Mirella Lapata. 2016. Long short-term memory-networks for machine reading. In *Proceedings of the Conference on Empirical Methods in Natural Language Processing.*

[10] Gardner, M., Grus, J., Neumann, M., Tafjord, O., Dasigi, P., Liu, N., Peters, M., Schmitz, M. and Zettlemoyer, L. 2018. AllenNLP: A deep semantic natural language processing platform. arXiv preprint arXiv:1803.07640.

[11] Jonas Gehring, Michael Auli, David Grangier, Denis Yarats, and Yann N Dauphin. 2017. Convolutional sequence to sequence learning. arXiv preprint arXiv:1705.03122.

[12] He, K., Zhang, X., Ren, S. and Sun, J., 2016. Deep residual learning for image recognition. In *Proceedings of the IEEE conference on computer vision and pattern recognition (pp. 770-778).*

[13] Sepp Hochreiter and Jürgen Schmidhuber. 1997. Long short-term memory. *Neural computation*, 9(8):1735–1780.

[14] Sepp Hochreiter, Yoshua Bengio, Paolo Frasconi, Jurgen Schmidhuber, et al. 2001. Gradient flow in recurrent nets: the difficulty of learning long-term dependencies.

[15] Urvashi Khandelwal, He He, Peng Qi, and Dan Jurafsky. 2018. Sharp nearby, fuzzy far away: How neurallanguage models use context. arXiv preprint arXiv:1805.04623.

[16] Guillaume Klein, Yoon Kim, Yuntian Deng, Jean Senellart, and Alexander M Rush. 2017. Opennmt: Open-source toolkit for neural machine translation. arXiv preprint arXiv:1701.02810.

[17] A. Krizhevsky, I. Sutskever, and G. Hinton. 2012. Imagenet classification with deep convolutional neural networks. In *Advances in neural information processing systems (NIPS).*

[18] Kenton Lee, Omer Levy, and Luke Zettlemoyer. 2017. Recurrent additive networks. CoRR abs/1705.07393.

[19] Marcus, Marcinkiewicz and Santorini. 1993. appeared in *Computational Linguistics*, June 1993 issue 19(2), pages 313-330.

[20] Lin, Chin-Yew. 2004. ROUGE: a Package for Auto-matic Evaluation of Summaries. In *Proceedings of the Workshop on Text Summarization Branches Out (WAS 2004),* Barcelona, Spain, July 25 - 26, 2004.

[21] Niehues, J. and Cho, E. 2017. Exploiting Linguistic Resources for Neural Machine Translation Using Multi-task Learning. In *Proceedings of WMT-2017,* pages 80–89, Copenhagen, Denmark.

[22] Jacob Devlin, Ming-Wei Chang, Kenton Lee, and Kristina Toutanova. 2018. Bert: Pre-training of deepbidirectional transformers for language understanding. arXiv preprint arXiv:1810.04805.

[23] Sennrich, R. and Haddow, B. 2016. Linguistic input features improve neural machine translation. In *Proceedings of the First Conference on Machine Translation,* pages 83–91, Berlin, Germany.

[24] Kyunghyun Cho, Bart van Merrienboer, Çaglar Gülçehre, Dzmitry Bahdanau, Fethi Bougares, Holger Schwenk, and Yoshua Bengio. 2014. Learning phrase representations using RNN encoder-decoder for statistical machine translation. In *Proceedings of EMNLP*, pages 1724–1734.

[25] Tomas Mikolov, Ilya Sutskever, Kai Chen, Greg S Corrado, and Jeff Dean. 2013. Distributed representations of words and phrases and their compositionality. In *Advances in Neural Information Processing Systems,* pages 3111-3119.

[26] Jeffrey Pennington, Richard Socher, and Christopher D. Manning. 2014. Glove: Global vectors for word 189 representation. In *Proceedings of EMNLP,* pages 1532-1543.

[27] Kishore Papineni, Salim Roukos, Todd Ward, and Wei-Jing Zhu. 2002. BLEU: a method for automatic evaluation of machine translation. In *Proceedings of Association for Computational Linguistics (ACL).* 2002.

[28] Alec Radford, Karthik Narasimhan, Tim Salimans, and Ilya Sutskever. 2018. Improving language understanding by generative pre-training. Technical report, OpenAI, 2018

[29] Ilya Sutskever, Oriol Vinyals, and Quoc V Le. 2014. Sequence to sequence learning with neural networks. In *Advances in neural information processing systems,* pages 3104–3112.

[30] Ashish Vaswani, Noam Shazeer, Niki Parmar, Jakob Uszkoreit, Llion Jones, Aidan N Gomez, Lukasz Kaiser, and Illia Polosukhin. 2017. Attention is all you need. In *Advances in Neural Information Processing Systems,* pages 6000–6010.

[31] Peter Shaw, Jakob Uszkoreit, and Ashish Vaswani. 2018. Self-attention with relative position representations. In *Proceedings of the North American Chapter of the Association for Computational Linguistics. (NAACL)*

[32] R. J. Williams and D. Zipser. 1989. A learning algorithm for continually running fully recurrent neural networks, Neural computation, vol. 1, no. 2, pp. 270-280.

[33] Yen-Chun Chen and Mohit Bansal. 2018. Fast abstractive summarization with reinforce-selected sentence rewriting. In *Proceedings of ACL-2018.*

[34] Zhenghua Li, Min Zhang, Wanxiang Che, Ting Liu, Wenliang Chen, and Haizhou Li. 2011. Joint Models for Chinese POS Tagging and Dependency Parsing. In *Proceedings of EMNLP.* pages 1180–1191.

[35] Ming Chen, Yingming Li, Zhongfei Zhang, and Siyu Huang.Tvt: Two-view transformer network for video captioning. In *Proceedings of Asian Conference on Machine Learning (ACML)*, 2018.

[36] Luowei Zhou, Yingbo Zhou, Jason J Corso, Richard Socher,and Caiming Xiong. End-to-end dense video captioning with masked transformer. In *Proceedings of the IEEE Conference on Computer Vision and Pattern Recognition (CVPR)*, pages 8739–8748, 2018.

[37] Xinxin Zhu, Lixiang Li, Jing Liu, Haipeng Peng,and Xinxin Niu. 2018. Captioning transformer with stacked attention modules. *Applied Sciences*, vol. 8, no. 5, pp. 739, 2018.

[38] Li, Jiangyun, Peng Yao, Longteng Guo, and Weicun Zhang. Boosted Transformer for Image Captioning. *Applied Sciences* 9, no.16 (2019):3260.